\documentclass[wcp]{jmlr}

\usepackage{longtable}
\usepackage{booktabs}

\usepackage[]{algorithm2e}
\usepackage{xcolor}

\newcommand{\hugo}[1]{{#1}}
\definecolor{OliveGreen}{cmyk}{0.64,0,0.95,0.40}

\DeclareMathOperator*{\argmax}{arg\,max}
\DeclareMathOperator*{\argmin}{arg\,min}
\newcommand{\ie}{\textit{i.e., }}
\newcommand{\eg}{\textit{e.g., }}

\newcommand{\dist}{p}
\newcommand{\cum}{F}
\newcommand{\dec}{G}
\newcommand{\quant}{q}
\newcommand{\lquant}{\underline\quant}
\newcommand{\uquant}{\overline\quant}
\usepackage{mathtools}

\newtheorem{ex}{Example}

\newtheorem{lem}{Lemma}

\jmlrvolume{60}
\jmlryear{2016}
\jmlrworkshop{ACML 2016}

\title[Quantile RL]{Quantile Reinforcement Learning}

  \author{\Name{Hugo Gilbert} \Email{hugo.gilbert@lip6.fr}\\
  \addr Sorbonnes Universit\'es\\
   UPMC Univ Paris 06\\ 
   CNRS, LIP6 UMR 7606
   \\ Paris, France
  \AND
  \Name{Paul Weng} \Email{paweng@cmu.edu}\\
  \addr School of Electronics and Information Technology\\
       SYSU-CMU Joint Institute of Engineering\\
       SYSU-CMU Shunde Joint Research Institute\\
       Guangzhou, 510006, PR China
 }

\begin{document}

\maketitle

\begin{abstract}
In reinforcement learning, the standard criterion to evaluate policies in a state is the expectation of (discounted) sum of rewards.
However, this criterion may not always be suitable, we consider an alternative criterion based on the notion of quantiles.
In the case of episodic reinforcement learning problems, we propose an algorithm based on stochastic approximation with two timescales.
We evaluate our proposition on a simple model of the TV show, Who wants to be a millionaire.
\end{abstract}

\section{Introduction}\label{sec:intro}

Markov decision process and reinforcement learning are powerful frameworks for building autonomous agents (physical or virtual), which are systems that make decisions without human supervision in order to perform a given task.
Examples of such systems abound: expert backgammon player \citep{Tesauro95}, dialogue systems \citep{ZhangCaiMaoChangGuo01}, 
acrobatic helicopter flight \citep{AbbeelCoatesNg10} or human-level video game player \citep{MnihKavukcuogluSilverRusuVenessBellemareGravesRiedmillerFidjelandOstrovskiPetersenBeattieSadikAntonoglouKingKumaranWierstraLeggHassabis15}.
However, the standard framework assumes that rewards are numeric, scalar and additive and that policies are evaluated with the expectation criterion.
In practice, it may happen that such numerical rewards are not available, for instance, when the agent interacts with a human who generally gives ordinal feedback (\eg ``excellent", ``good", ``bad" and so on).
Besides, even when this numerical information is available, one may want to optimize a criterion different than the expectation, for instance in one-shot decision-making.

Several works considered the case where preferences are qualitative.
Markov decision processes with ordinal reward have been investigated \citep{Weng12,Weng11} and different ordinal decision criteria have been proposed in that context. 
More generally, preference-based reinforcement learning \citep{AkrourSchoenauerSebag12,FurnkranzHullermeierChengPark12,BusaFeketeSzorenyiWengChengHullermeier13EWRL,BusaFeketeSzorenyiWengChengHullermeier14} has been proposed to tackle situations where the only available preferential information concerns pairwise comparisons of histories.

In this paper, we propose to search for a policy that optimizes a quantile instead of the expectation.
Intuitively, the $\tau$-quantile of a distribution is the value $q$ such that the probability of getting a value lower than $q$ is $\tau$ (and therefore the probability of getting a value greater than $q$ is $1-\tau$).
The median is an example of quantile where $\tau = 0.5$.
Interestingly, in order to use this criterion only an order over valuations is needed.

The quantile criterion is extensively used as a decision criterion in many domains.
In finance, it is a risk measure and is known as Value-at-Risk \citep{Jorion06}.
For its cloud computing services, Amazon reports \citep{DeCandiaHastorunJampaniKakulapatiLakshmanPilchinSivasubramanianVosshallVogels07} that they optimize the 99.9\%-quantile\footnote{Or 0.01\%-quantile, depending on whether the problem is expressed in terms of costs or rewards.}.
In fact, decisions in the web industry are often made based on quantiles \citep{WolskiBrevik14,DeCandiaHastorunJampaniKakulapatiLakshmanPilchinSivasubramanianVosshallVogels07}. 
More generally, in the service industry, because of skewed distributions \citep{BenoitVandenPoel09}, one generally does not want that customers are satisfied on average, but rather that most customers (\eg 99\% of them) to be as satisfied as possible.

The use of the quantile criterion can be explained by the nice properties it enjoys: 
\begin{itemize}
\item preferences and uncertainty can be valued on scales that are not commensurable, 
\item preferences over actions or trajectories can be expressed on a purely ordinal scale, 
\item preferences over policies are more robust than with the standard criterion of maximizing the expectation of cumulated rewards.
\end{itemize}

The contributions of this paper are as follows.
To the best of our knowledge, we are the first to propose an RL algorithm to learn a policy optimal for the quantile criterion.
This algorithm is based on stochastic approximation with two timescales. 
We present an empirical evaluation of our proposition on a version of Who wants to be a millionaire.

The paper is organized as follows.
Section~\ref{sec:related} presents the related work.
Section~\ref{sec:back} recalls the necessary background for presenting our approach.
Section~\ref{sec:qrl} states the problem we solve and introduce our algorithm, Quantile Q-learning.
Section~\ref{sec:expe} presents some experimental results.
Finally, we conclude in Section~\ref{sec:conclu}.

\section{Related Work}\label{sec:related}

A great deal of research on MDPs \citep{BBMSWeng10} considered decision criteria different to the standard ones (i.e., expected discounted sum of rewards, expected total rewards or expected average rewards).
For instance, in the operations research community, 
\cite{White87} notably considered different cases where preferences over policies only depend on sums of rewards: Expected Utility (EU), probabilistic constraints and mean-variance formulations.
In this context, he showed the sufficiency of working in a state space augmented with the sum of rewards obtained so far.
\cite{FilarKallenbergLee89} investigated decision criteria that are variance-penalized versions of the standard ones.
They formulated the obtained optimization problem as a non-linear program.
\cite{Yu98} optimized the probability that the total reward becomes higher than a certain threshold. 


Additionally, in the artificial intelligence community, \cite{LiuKoenig05,LiuKoenig06} also investigated the use of EU as a decision criterion in MDPs. 
To optimize it, they proposed a functional variation of Value Iteration. 
In the continuation of this work, \cite{GilbertSpanjaardViappianiWeng15} investigated the use of Skew-Symmetric Bilinear (SSB) utility \citep{Fishburn81} functions --- a generalization of EU that enables intransitive behaviors and violation of the independence axiom --- as decision criteria in finite-horizon MDPs. 
Interestingly, SSB also encompasses probabilistic dominance, a decision criterion that is employed in preference-based sequential decision-making \citep{AkrourSchoenauerSebag12,FurnkranzHullermeierChengPark12,BusaFeketeSzorenyiWengChengHullermeier13EWRL,BusaFeketeSzorenyiWengChengHullermeier14}.

In theoretical computer science, sophisticated decision criteria have also been studied in MDPs.
For instance, \cite{Gimbert07} proved that many decision criteria based on expectation (of limsup, parity... of rewards) admit a stationary deterministic optimal policy.
\cite{BruyereFiliotRandourRaskin14} considered sophisticated preferences over policies, which amounts to searching for policies that maximize the standard criterion while ensuring an expected sum of rewards higher than a threshold with probability higher than a fixed value.
This work has also been extended to the multiobjective setting \citep{RandourRaskinSankur14}.

Recent work in Markov decision process and reinforcement learning considered conditional Value-at-risk (CVaR), a criterion related to quantile, as a risk measure.
\cite{BauerleOtt11} proved the existence of deterministic wealth-Markovian policies optimal with respect to CVaR. 
\cite{ChowGhavamzadeh14} proposed gradient-based algorithms for CVaR optimization.
In contrast, \cite{BorkarJain14} used CVaR in constraints instead of as objective function.

Closer to our work, several quantile-based decision models have been investigated in different contexts. 
In uncertain MDPs where the parameters of the transition and reward functions are imprecisely known, \cite{DelageMannor07} presented and investigated a quantile-like criterion to capture the trade-off between optimistic and pessimistic viewpoints  on an uncertain MDP. 
The quantile criterion they use is 
different to ours as it takes into account the uncertainty present in the parameters of the MDP.

In MDPs with ordinal rewards \citep{Weng11,Weng12,Filar83}, quantile-based decision models were proposed to compute policies that maximize a quantile using linear programming.
While quantiles in those works are defined on distributions over ordinal rewards, quantiles in this paper are defined on distributions over histories. 

More recently, in the machine learning community, quantile-based criteria have been proposed in the multi-armed bandit (MAB) setting, a special case of reinforcement learning.
\cite{YuNikolova13} proposed an algorithm in the pure exploration setting for different risk measures, including Value-at-Risk. 
\cite{CarpentierValko14} studied the problem of identifying arms with extreme payoffs, a particular case of quantiles.
Finally, \cite{SzorenyiBusa-FeketeWengHullermeier15} investigated MAB problems where a quantile is optimized instead of the mean.

The algorithm we propose is based on stochastic approximation with two timescales, a technique proposed by \cite{Borkar97,Borkar08}.
This method has recently been exploited in achievability problems \citep{Blackwell56} in the context of multiobjective MDPs \citep{KalathilBorkarJain14} and for learning SSB-optimal policies \citep{GilbertZanuttiniViappianiWengNicart16}.

\section{Background}\label{sec:back}

We provide in this section the background information necessary to present our algorithm  to learn a policy optimal for the quantile criterion.

\subsection{Markov Decision Process}

\textit{Markov Decision Processes} (MDPs) offer a powerful formalism to model and solve sequential decision-making problems \citep{Puterman94}.
A finite horizon MDP is formally defined as a tuple $\mathcal{M}_T=(S, A, \mathbf{P}, \mathbf R, s_0)$ where:
\begin{itemize} 
\item $T$ is a finite horizon, 
\item $S$ is a finite set of states, 
\item $A$ is a finite set of actions, 
\item $\mathbf{P} : S \times A \times S \rightarrow \mathbb{R}$ is a transition function with $\mathcal{P}(s,a,s')$ being the probability of reaching state $s'$ when action $a$ is performed in state $s$, 
\item $\mathbf R : S \times A \rightarrow \mathbb{R}$ is a bounded reward function and 
\item $s_0 \in S$ is an initial state. 
\end{itemize}

In this model, starting from initial state $s_0$, an agent chooses at every time step $t$ an action $a_t$ to perform in her current state $s_t$, which she can observe.
This action results in a new state $s_{t+1} \in S$ according to probability distribution $\mathbf{P}(s_t, a_t, .)$, and a reward signal $\mathbf R(s_t, a_t)$, which penalizes or reinforces the choice of this action. 

We will call {\em $t$-history} $h_t$ a succession of $t$ state-action pairs starting from state $s_0$ (\eg $h_t = (s_0,a_1,s_1,\ldots,s_{t-1},a_{t-1},s_t)$). 
The action choices of the agent is guided by a policy.
More formally, a {\em policy} $\pi$ at an horizon $T$ is a sequence of $T$ decision rules $(\delta_T,\ldots,\delta_{1})$. 
{\em Decision rules} prescribe which action the agent should perform at a given time step. 
They can be \textit{Markovian} if they only depend on the current state. 
Besides, a decision rule is either {\em deterministic} if it always selects the same action in a given situation or {\em randomized} if it prescribes a probability distribution over possible actions. 
Consequently, a policy can be \textit{Markovian}, \textit{deterministic} or  \textit{randomized} according to the type of its decision rules. 
Lastly, a policy is \textit{stationary} if it applies the same decision rule at every time step, \ie $\pi = (\delta,\delta,\ldots)$.

Policies can be compared with respect to different decision criteria.
The usual criterion is the {\em expected (discounted) sum of rewards}, for which an optimal deterministic Markovian policy is known to exist for any horizon $T$.
This criterion is defined as follows.
First, the value of a history $h_t = (s_0,a_1,s_1,\ldots,s_{t-1},a_{t},s_t)$ is described as the (possibly discounted) sum of rewards obtained along it, \ie
\begin{align*}
r(h_t) = \sum_{i=1}^{t} \gamma^{i-1} \mathbf R(s_{i-1}, a_{i})
\end{align*}
where $\gamma \in [0, 1]$ is a discount factor.
Then, the value of a policy $\pi = (\delta_T, \ldots, \delta_1)$ in a state $s$ is set to be the expected value of the histories that can be generated by $\pi$ from $s$.
This value, given by the {\em value function} $v^\pi_T : S \to \mathbb R$ can be computed iteratively as follows:
\begin{align}
      v^{\pi}_{0}(s)&= 0 \nonumber\\ 
v^{\pi}_t(s)&=  \mathbf R(s,\delta_t(s)) + \gamma \sum_{s'\in S}\mathbf{P}(s,\delta_t(s),s')v^{\pi}_{t-1}(s')\label{eq:value_function}
\end{align}




The value $v^{\pi}_t(s)$ is the expectation of cumulated rewards obtained by the agent if she performs action $\delta_t(s)$ in state $s$ at time-step $t$ and continues to follow policy $\pi$ thereafter. 
The higher the values of $v^{\pi}_t(s)$ are, the better. 
Therefore, value functions induce a preference relation $\succsim_{\pi}$ over policies in the following way:
\begin{align*}
\pi \succsim_{\pi} \pi' \Leftrightarrow \forall s\in S, v^{\pi}_T(s) \geq v^{\pi'}_T(s)
\end{align*}

A solution to an MDP is a policy, called {\em optimal policy}, that ranks the highest with respect to $\succsim_{\pi}$. Such a policy can be found by solving the 
following equations, which yields the value function of an optimal policy:
\begin{align*}
      v^{*}_{0}(s)&= 0 \\ 
      v^{*}_t(s)&=\max_{a\in A}  \mathbf R(s,a)+ \gamma \sum_{s'\in S}\mathbf{P}(s,a,s')v^{*}_{t-1}(s')
\label{eq:bellman}
\end{align*}

\subsection{Reinforcement Learning}

In the reinforcement learning (RL) setting, the assumption of the knowledge of the environment is relaxed: both dynamics through the transition function and preferences via the reward function are not known anymore.
While interacting with its environment, an RL agent tries to learn a good policy by trial and error.

To make finite horizon MDPs learnable, we assume the decision process is repeated infinitely many times. 
That is, when horizon $T$ is reached, 
we assume that the agent automatically returns to the initial state and the problem starts over.

A simple algorithm to solve such an RL problem is the Q-learning algorithm (see Algorithm~\ref{alg:qlearning}), which estimates the Q-function:
\begin{align*}
Q_t(s, a) = \mathbf R(s, a) + \gamma \sum_{s' \in S} \mathbf P(s, a, s') V_{t-1}(s')
\end{align*}
And obviously, we have:
$V_t(s) = \displaystyle\max_{a\in A} Q_t(s, a)$.

In Algorithm~\ref{alg:qlearning}, Line~\ref{algl:choice} generally depends on the $Q_{t-1}(s, \cdot)$ and possibly on iteration $n$.
A simple strategy to perform this choice is called {\em $\varepsilon$-greedy} where the best action dictated by $Q_{t-1}(s, \cdot)$ is chosen with probability $1 - \varepsilon$ (with $\varepsilon$ a small positive value) or a random action is chosen otherwise. 
A schedule can be defined so that parameter $\varepsilon$ tends to zero as $n$ tends to infinity.
Besides, $\alpha_n(s, a) \in (0, 1)$ on Line~\ref{algl:qupdate} is a learning rate. 
In the general case, it depends on iteration $n$, state $s$ and action $a$, although in practice it is often chosen as a constant.

\begin{algorithm}[bt]
\DontPrintSemicolon
\KwData{$\mathcal M_T = (S, A, G, \mathbf P, \mathbf R, s_0)$ MDP}
\KwResult{$Q$}
\Begin{
$Q_0(s, a) \longleftarrow 0, \forall (s, a)\in S\times A$\\
$s \longleftarrow s_0$\\
$t \longleftarrow 1$\\
\For{$n=1$ to $N$}{
   \nl $a \longleftarrow$ choose action\\ \label{algl:choice}
   $r, s' \longleftarrow$ perform action $a$ in state $s$\\
   \nl $Q_t(s, a) \longleftarrow Q_t(s, a) + \alpha_n(s, a)\big(r + \gamma \max_{a' \in A} Q_{t-1}(s', a') - Q_t(s, a)\big)$\label{algl:qupdate}\\ 
   \nl \uIf{$t = T$}{ \label{algl:if}
      $s \longleftarrow s_0$\\
      $t \longleftarrow 1$
   } \Else{
      $s \longleftarrow s'$\\
      $t \longleftarrow t  +1$
   }
}
}
\caption{Q-learning}\label{alg:qlearning}
\end{algorithm}

\subsection{Limits of standard criteria}

The standard decision criteria used in MDPs, which are based on expectation, may not be reasonable in some situations.
Firstly, unfortunately, in many cases, the reward function $\mathbf R$ is not known.
In those cases, one can try to recover the reward function from a human expert \citep{NgRussell00,ReganBoutilier09,WengZanuttini13}. 
However, even for an expert user, the elicitation of the reward function can reveal burdensome. 
In inverse reinforcement learning \citep{NgRussell00}, the expert is assumed to know an optimal policy, which is rarely true in practice.
In interactive settings \citep{ReganBoutilier09,WengZanuttini13}, this elicitation process can be cognitively very complex as it requires to balance several criteria in a complex manner and as it can imply a large number of parameters. 
In this paper, we address this problem by only assuming that a strict weak ordering over histories is known. 

Secondly, for numerous applications, the expectation of cumulated reward, as used in Equation~\ref{eq:value_function}, may not be the most appropriate criterion (even when a numeric reward function is defined). 
For instance, in case of high variance or when a policy is known to be only applied a few times, the solution given by this criterion may not be satisfying for risk-averse agent.
Moreover, in some domains (\eg web industry or more generally service industry), decisions about performance are often based on the minimal quality of $99\%$ of the possible outcomes. 
Therefore, in this article we aim at using a quantile (defined in Section~\ref{sec:quantile}) as a decision criterion to solve an MDP.

\subsection{MDP with End States}

In this paper, we work with episodic MDPs with end states.
Such an MDP is formally defined as a tuple $\mathcal{M}_T=(S, A, G, \mathbf{P}, s_0)$ where 
$S$, $A$, $\mathbf P$, $s_0$ are defined as previously, 
$G \subseteq S$ is a finite set of end states and
$T$ is a finite maximal horizon (i.e., an end state is attained after at most $T$ time steps.).
We call {\em episode} a history starting from $s_0$ and ending in a final state of $G$.

We assume that a preference relation is defined over end states: 
We write $g' \prec g$ if end state $g$ is preferred to end state $g'$.
Without loss of generality, we assume that $G = \{g_1, \ldots, g_n\}$ and end states are ordered with increasing preference, \ie 
$g_1 \prec g_2 \prec \ldots \prec g_n$.
The weak relation of $\prec$ is denoted $\preceq$.

Note that a finite horizon MDP can be reformulated as an MDP with end states by state augmentation.
Although the resulting MDP may have a large-sized state space, the two models are formally equivalent.
We focus on episodic MDPs with end states to simplify the presentation of our approach.

%

\subsection{Quantile Criterion}\label{sec:quantile}

We define quantiles of distributions over end states of $G$, which are ordered by $\preceq$.
Let $\tau \in [0, 1]$ be a fixed parameter.
Intuitively, the $\tau$-quantile of a distribution of end states, is the value $\quant \in G$ such that the probability of getting an end state equal or lower than $\quant$ is $\tau$ and that of getting an end state equal or greater than $\quant$ is $1-\tau$. 
The $0.5$-quantile, also known as median, can be seen as the ordinal counterpart of the mean.
The $0$-quantile (resp. $1$-quantile) is the minimum (resp. maximum) of a distribution.
More generally, quantiles, which have been axiomatically characterized by \cite{Rostek10}, define decision criteria that have the nice property of not requiring numeric valuations, but only an order.

The formal definition of quantiles can be stated as follows.
Let $\dist^{\pi}$ denote the probability distribution over end states induced by a policy $\pi$ from initial state $s_0$, 
the {\em cumulative distribution} induced by $\dist^\pi$ is then defined as $\cum^{\pi}$ where $\cum^{\pi}(g) = \sum_{g' \preceq g} \dist^{\pi}(g')$ is the probability of getting an end state not preferred to $g$ when applying policy $\pi$. 
Similarly, the {\em decumulative distribution} induced by $\dist^\pi$ is defined as $\dec^{\pi}(g) = \sum_{g \preceq g'} \dist^{\pi}(g')$ is the probability of getting an end state not lower than $g$.

These two notions of cumulative and decumulative enable us to define two kinds of criteria. 
First, given a policy $\pi$, we define the lower $\tau$-quantile for $\tau \in(0,1]$ as:
\begin{equation}
\lquant_{\tau}^{\pi} = \min\{g \in G \,|\, \cum^{\pi}(g) \geq \tau\} 
\end{equation}
where the $\min$ operator is with respect to $\preceq$.

Then, given a policy $\pi$, we define the upper $\tau$-quantile for $\tau \in[0,1)$ as:
\begin{equation}
\uquant_{\tau}^{\pi} = \max\{g \in G \,|\, \dec^{\pi}(g) \geq 1-\tau\} 
\end{equation}
where the $\max$ operator is with respect to $\preceq$.

If $\tau = 0$ or $\tau = 1$ only one of $\lquant_{\tau}^{\pi}$ or $\uquant_{\tau}^{\pi}$ is defined and we define the $\tau$-quantile $\quant_{\tau}^{\pi}$ as that value. 
When both are defined, by construction, we have $\lquant_{\tau}^{\pi} \preceq \uquant_{\tau}^{\pi}$.
If those two values are equal, $\quant_{\tau}^{\pi}$ is defined as equal to them. 
For instance, this is always the case in continuous settings for continuous distributions. 
However, in our discrete setting, it could happen that those values differ, as shown by Example~\ref{ex:1}. 
\begin{ex}\label{ex:1}
Consider an MDP where $G = \{g_1 \prec g_2 \prec g_3\}$. 
Let $\pi$ be a policy that attains each end state with probabilities $0.5$, $0.2$ and $0.3$ respectively. 
It is easy to check that  $\lquant_{0.5}^{\pi}  = g_1$ whereas $\uquant_{0.5}^{\pi} = g_2$.
\end{ex}
When the lower and the upper quantiles differ, one may define the quantile as a function of the lower and upper quantiles \citep{Weng12}.
For simplicity, in this paper, we focus on optimizing directly the lower and the upper quantiles.

The quantile criterion is difficult to optimize, even when a numerical reward function is given and the quality of an episode is defined as the cumulative of rewards received along the episode.
This difficulty comes notably from two related sources:
\begin{itemize}
\item The quantile criterion is {\em non-linear}: for instance, the $\tau$-quantile $\quant_{\tau}^{\tilde{\pi}}$ of the mixed policy $\tilde{\pi}$ that generates an episode using policy $\pi$ with probability $p$ and $\pi'$ with probability $1-p$ is not equal to $p \quant_{\tau}^{\pi} + (1-p)\quant_{\tau}^{\pi'}$.
\item The quantile criterion is {\em non-dynamically consistent}: 
A sub-policy at time step $t$ of an optimal policy for horizon $T$ may not be optimal for horizon $T-t$.
\end{itemize}
In decision theory \citep{McClennen90}, three approaches have been considered for such kinds of decision criteria: 
\begin{enumerate}
\item {\em Consequentialist} approach: at each time step $t$, follow an optimal policy for the problem with horizon $T-t$ and initial state $s_t$ even if the resulting policy is not optimal at horizon $T$; 
\item {\em Resolute choice} approach: at time step $t=0$, apply an optimal policy for the problem with horizon $T$ and initial state $s_0$ and do not deviate from it;
\item {\em Sophisticated resolute choice} approach \citep{Jaffray98,FargierJeantetSpanjaard11}: apply a policy $\pi$ (chosen at the beginning) that trades off between how much $\pi$ is optimal for all horizons $T, T-1, \ldots, 1$.
\end{enumerate}

With non-dynamically consistent preferences, it is debatable to adopt a consequentialist approach, as the sequence of decisions may lead to dominated results. 
In this paper, we adopt a resolute choice point of view.
We leave the third approach for future work.

\section{Quantile-based Reinforcement Learning}\label{sec:qrl}

In this section, we first state the problem solved in this paper and some useful properties.
Then, we present our algorithm called Quantile Q-learning (or QQ-learning for short), which is an extension of Q-learning and exploits a two-timescale stochastic approximation technique.

\subsection{Problem Statement}

In this paper, we aim at learning a policy that is optimal for the quantile criterion from a fixed initial state.
We assume that the underlying MDP is an episodic MDP with end states.
Let $\tau \in (0,1)$ be a fixed parameter.
Formally, the problem of determining a policy optimal for the lower/upper $\tau$-quantile can be stated as follows:
\begin{align}\label{eq:pb}
\pi^* = \argmax_\pi \lquant^\pi_\tau \quad \mbox{ or }\quad \pi^* = \argmax_\pi \uquant^\pi_\tau
\end{align}
We focus on learning a policy that is 
deterministic and Markovian.


%
The optimal lower/upper quantiles satisfy the following lemmas:
\begin{lem}
The optimal lower $\tau$-quantile $\lquant^*_\tau$ satisfies:
\begin{align}
\lquant^*_\tau &= \min\{g : \cum^{*}(g) \geq \tau \} \label{eq:bslq1}\\
\cum^{*}(g) &= \min_\pi \cum^{\pi}(g) \quad \forall g \in G \label{eq:pblq1}
\end{align}
and the optimal upper $\tau$-quantile $\uquant^*_\tau$ satisfies:
\begin{align}
\uquant^*_\tau &= \max\{g : \dec^{*}(g) \geq 1 - \tau \} \label{eq:bsuq1}\\
\dec^{*}(g) &= \max_\pi \dec^{\pi}(g)  \quad \forall g \in G \label{eq:pbuq1}
\end{align}
\label{lem:4}
\end{lem}
%

Then, if the optimal lower quantile ($\lquant^*_\tau$) or upper quantile ($\uquant^*_\tau$) were known, the problem would be relatively easy to solve.
By Lemma~\ref{lem:4}, an optimal policy for the lower quantile could be obtained as follows:
\begin{align}\label{eq:pblq}
\pi^* = \argmin_\pi \cum^\pi(\lquant^-_\tau)
\end{align}
where $\lquant^-_\tau = g_1$ if $\lquant^*_\tau = g_1$ and $\lquant^-_\tau = g_i$ if $\lquant^*_\tau = g_{i+1}$.
The reason one needs to use $\lquant^-_\tau$ instead of the optimal lower quantile is that otherwise it may happen that the cumulative distribution of a non-optimal policy $\pi$ is 
smaller than or equal to that of 
an optimal policy at the lower quantile $\lquant^*_\tau$ (but greater below $\lquant^*_\tau$ as $\pi$ is not optimal). 
In that case, the lower quantile of $\pi$ might not be $\lquant^*_\tau$.
Such a thing cannot happen for the upper quantile. 
In that sense, determining an optimal policy for the upper quantile is easier.
By Lemma~\ref{lem:4}, it is simply given by:
\begin{align}\label{eq:pbuq}
\pi^* = \argmax_\pi \dec^\pi(\uquant^*_\tau)
\end{align}
In practice, if the lower/upper quantiles were known, those policies could be computed by solving a standard MDP with the following reward functions:
\begin{align*}
\underline {\mathbf R}_{\theta}(s) = \left\{
\begin{array}{ll} 
0 & \forall s \not\in \mathcal{G} \\
0 & \forall s = g_i \in \mathcal{G}, \theta \geq i\\
1 & \forall s = g_i \in \mathcal{G}, \theta \leq i - 1
\end{array}\right.
\end{align*}
for the lower quantile with $\theta = k$ if $\lquant^-_\tau = g_k$ and
\begin{align*}
\overline {\mathbf R}_{\theta}(s) = \left\{
\begin{array}{ll} 
0 & \forall s \not\in \mathcal{G} \\
1 & \forall s = g_i \in \mathcal{G}, \theta \leq i\\
0 & \forall s = g_i \in \mathcal{G}, \theta \geq i + 1
\end{array}\right.
\end{align*}
for the upper quantile with $\theta = k$ if $\uquant^*_\tau = g_k$.
Note that $\underline {\mathbf R}_{\theta}$ can be rewritten to depend on the optimal lower quantile:
\begin{align*}
\underline {\mathbf R}_{\theta}(s) = \left\{
\begin{array}{ll} 
0 & \forall s \not\in \mathcal{G} \\
0 & \forall s = g_i \in \mathcal{G}, \theta \geq i+1\\
1 & \forall s = g_i \in \mathcal{G}, \theta \leq i 
\end{array}\right.
\end{align*}
with $\theta = k$ if $\lquant^*_\tau = g_k$.

Solving an MDP with $\underline {\mathbf R}_{\theta}$ amounts to minimizing the probability of ending in a final state strictly less preferred than $\lquant^*_\tau$, which solves Equation~\ref{eq:pblq}.
Similarly, solving an MDP with $\overline {\mathbf R}_{\theta}$ amounts to maximizing the probability of ending in a final state at least as preferred as $\uquant^*_\tau$, which solves Equation~\ref{eq:pbuq}.

Now, the issue here is that the lower and upper quantiles are not known.
We show in the next subsection that this problem can be overcome with a two-timescale stochastic approximation technique.

\subsection{QQ-learning}\label{sec:algo}


As the lower and upper quantiles are not known, we let parameter $\theta$ vary in $\mathbb{R}^+$ during the learning steps and we refine the definition of the previous reward functions to make sure they are both well-defined for all $\theta \in \mathbb{R}^+$ and smooth in $\theta$:
\begin{align*}
\underline{\mathbf R}_{\theta}(s) = \left\{
\begin{array}{ll} 
0 & \forall s \not\in \mathcal{G} \\
-1 & \forall s = g_i \in \mathcal{G}, \theta \geq i +1\\
0 & \forall s = g_i \in \mathcal{G}, \theta \leq i \\
i- \theta & \text{ else }
\end{array}\right.
\end{align*}
\begin{align*}
\overline{\mathbf R}_{\theta}(s) = \left\{
\begin{array}{ll} 
0 & \forall s \not\in \mathcal{G} \\
1 & \forall s = g_i \in \mathcal{G}, \theta \leq i\\
0 & \forall s = g_i \in \mathcal{G}, \theta \geq i + 1\\
i+1- \theta & \text{ else }
\end{array}\right.
\end{align*}


In the remaining of the paper, we present how to solve for the upper quantile. 
A similar approach can be developed for the lower quantile.
In order to find the optimal upper quantile, one could use the strategy described in Algorithm~\ref{alg:search}.
Value $V^*_{\theta}(s_0)$ approximates the probability of reaching an end state whose index is at least as high as $\theta$.
If that value is smaller than $1-\tau$, it means $\theta$ is too high and should be decreased.
Otherwise $\theta$ is too small and should be increased.
Parameter $\theta$ will then converge to the index of the optimal upper quantile, which is the maximal value for $\theta$ such that $V^*_{\theta}(s_0) \ge 1 - \tau$.
The optimal policy for $V^*_\theta$ is an optimal policy for the upper $\tau$-quantile. 

\begin{algorithm}[tb!]
\DontPrintSemicolon
\KwData{$\mathcal{M}_T=(S, A, G, \mathbf{P}, s_0)$}
\Begin{
Initialize $\theta$ to a random value\\
\For{$n=1, 2, \ldots $}{
   \nl Solve MDP $\mathcal{M}_T$ with reward function $\overline{\mathbf{R}}_{\theta}$\; \label{algl:solve}\\
   \uIf{$V^*_{\theta}(s_0) < 1 - \tau$}{
      $\theta \longleftarrow \theta - 1/n$\;
   } \Else{
      $\theta \longleftarrow \theta + 1/n$\;
   }
}
}
\caption{Simple strategy for finding the optimal upper quantile}\label{alg:search}
\end{algorithm}

In a reinforcement learning setting, the solve MDP part (line~\ref{algl:solve} in Algorithm~\ref{alg:search}) could be replaced by an RL algorithm such as Q-learning. 
The problem is that such algorithm is only guaranteed to converge to the solution when $n \to \infty$. 
It would be therefore difficult to integrate Q-learning in Algorithm~\ref{alg:search}.
Instead, a good policy can be learned while searching for the correct value of $\theta$.
To that aim, we use a two-timescale technique \citep{Borkar97,Borkar08} in which Q-learning and the update of parameter $\theta$ are run concurrently but at different speeds (\ie at two different timescales). 
For this to work, parameter $\theta$ needs to be seen as quasi-static for the Q-learning algorithm.
This is possible if the ratio of the learning rate of Q-learning and that of the update of $\theta$ satisfies:
\begin{align}\label{eq:ratio}
\lim_{n \to \infty} \frac{\beta_n}{\alpha_n} =0
\end{align}
where $\beta_n = 1/n$ is the learning rate for parameter $\theta$ and $\alpha_n$ is the learning rate in the Q-learning algorithm.
Equation~\ref{eq:ratio} implies that parameter $\theta$ is changing at a slower timescale than the $Q$ function.


\begin{algorithm}[tb!]
\DontPrintSemicolon
\KwData{$\mathcal{M}_T=(S, A, G, \mathbf{P}, s_0)$ with $\mathbf P$ unknown}
\KwResult{$Q$}
\Begin{
$Q(s, a) \longleftarrow 0, \forall (s, a)\in S\times A$\\
$s \longleftarrow s_0$\\
$t \longleftarrow 1$\\
\For{$n=1$ to $N$}{
   $a \longleftarrow$ choose action\\
   $r, s' \longleftarrow$ perform action $a$ in state $s$\\
   $Q_t(s, a) \longleftarrow Q_t(s, a) + \alpha_n(s, a)\big(r + \gamma \max_{a' \in A} Q_{t-1}(s', a') - Q_t(s, a)\big)$\\
   \uIf{$V^*_{\theta}(s_0) < 1 - \tau$}{
      $\theta \longleftarrow \theta - 1/n$
   } \Else{
      $\theta \longleftarrow \theta + 1/n$
   }
   \uIf{$s' \in G$}{
      $s \longleftarrow s_0$\\
      $t \longleftarrow 1$
   } \Else{
      $s \longleftarrow s'$\\
      $t \longleftarrow t  +1$
   }
}
}
\caption{QQ-learning for the upper-quantile}\label{alg:qqlearning}
\end{algorithm}

\section{Experimental Results}\label{sec:expe}

To demonstrate the soundness of our approach, we evaluate our algorithm on the domain, Who wants to be a millionaire.
We present the experimental results below.

\subsection{Domain}

In this popular television game show, a contestant needs to answer a maximum of 15 multiple-choice questions (with four possible answers) of increasing difficulty, for increasingly large sums, roughly doubling the pot at each question. At each time step, the contestant may decide to walk away with the money currently won. If she answers incorrectly, then all winnings are lost except what has been earned at a ``guarantee point'' (questions 5 and 10). The player is allowed 3 lifelines (50:50, which removes two choices, ask the audience and call a friend for suggestions); each can only be used once. We used the first model of the Spanish 2003 version of the game presented by \cite{PereaPuerto07}. The probability of answering correctly is a function of the question's number and increased by the lifelines used (if any).

\subsection{Results}

\begin{figure}
  \centering
  \subfigure[]{\label{fig:exp}\scalebox{0.9}{\includegraphics[height = 6cm]{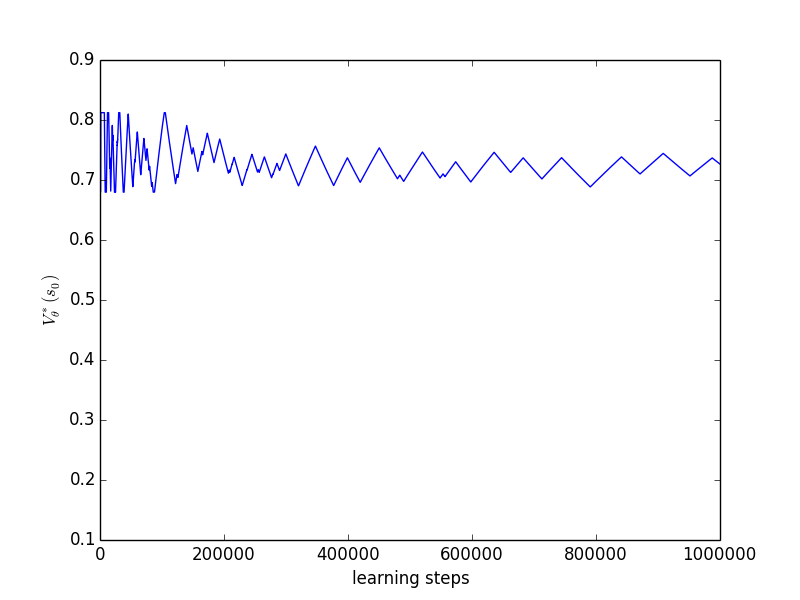}}}
  \hspace{5pt}
 \subfigure[]{\label{fig:exp2}\scalebox{0.9}{\includegraphics[height = 6cm]{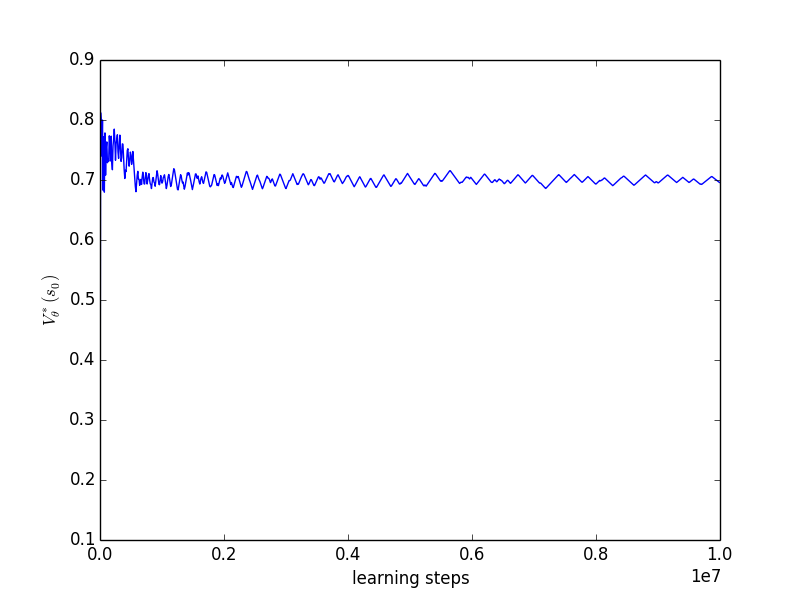}}}
  \hspace{5pt}
  \subfigure[]{\label{fig:exp3}\scalebox{0.9}{\includegraphics[height = 6cm]{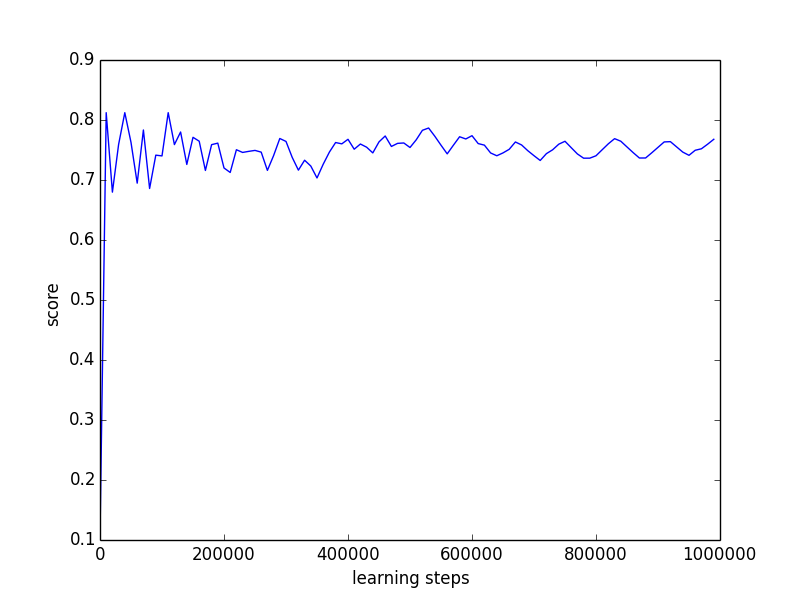} }}
  \hspace{5pt}
  \subfigure[]{\label{fig:exp4}\scalebox{0.9}{\includegraphics[height = 6cm]{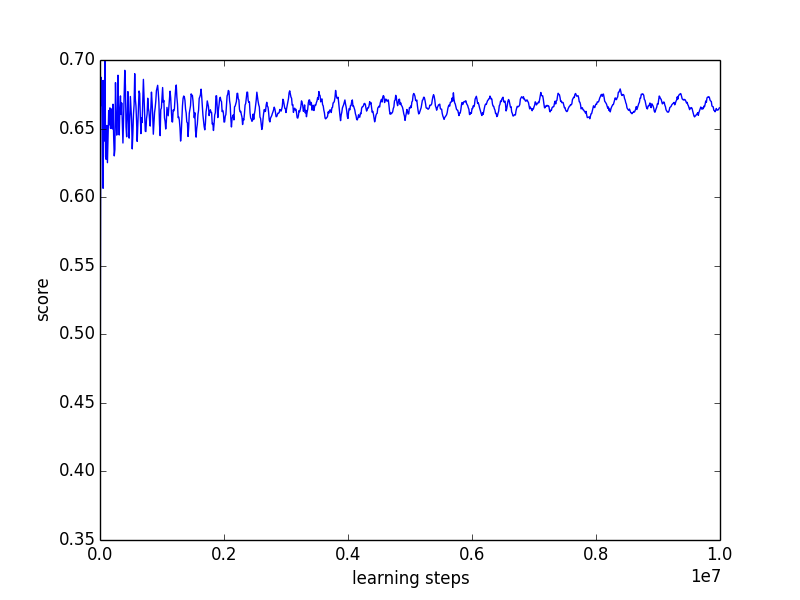}}}
  \hspace{5pt}
  \subfigure[]{\label{fig:exp5}\scalebox{0.9}{\includegraphics[height = 6cm]{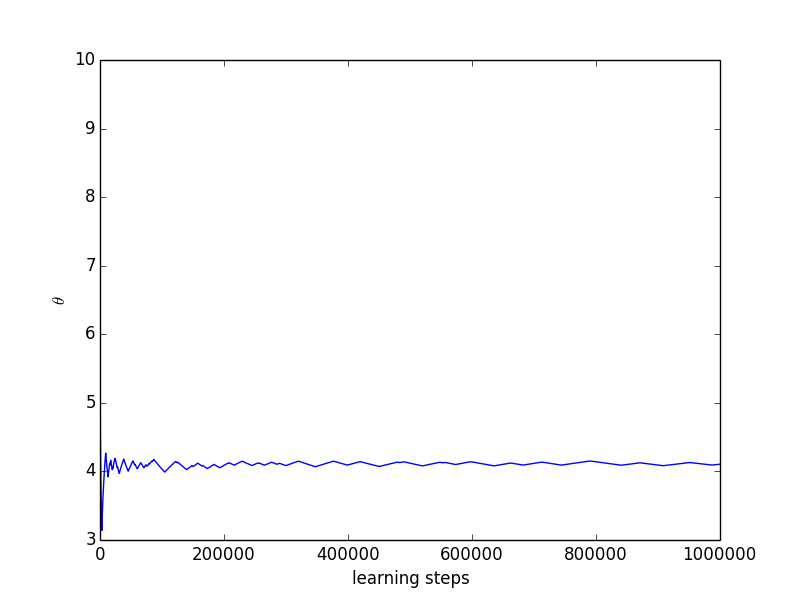} }}
  \hspace{5pt}
  \subfigure[]{\label{fig:exp6}\scalebox{0.9}{\includegraphics[height = 6cm]{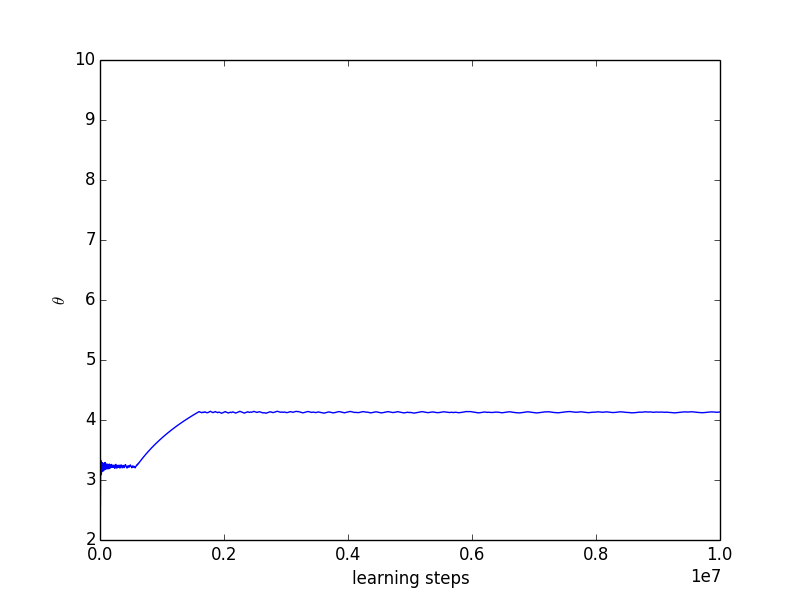}}}
\caption{Evolution of $V^*_\theta(s_0)$, score and $\theta$ for 1 million learning steps (left) and 10 million learning steps (right)}\label{fig:res}
 \end{figure}

We plot the results (see Figure~\ref{fig:res}) obtained for two different learning runs, one with 1 million learning steps and the other with 10 million learning steps. The $\theta$-updates were interleaved with a Q-learning phase using an $\varepsilon$-greedy exploration strategy with $\varepsilon = 0.01$ and with learning rate $\alpha(n) = 1/(n+1)^{11/20}$. 
One can check that Equation~\ref{eq:ratio} is satisfied.
\hugo{ We optimize the upper quantile with $\tau = 0.3$.  During the learning process we maintain a vector $\mathbf f$ of frequencies with which each final state has been attained. We define the quantity {\em score} as the cross product between $\mathbf f$ and the vector of rewards obtained when attaining each final state given the current value of $\theta$. Put another way, {\em score} is the value of the non-stationary policy that has been played since the beginning of the learning process. Moreover, at each iteration we compute $V_\theta^*(s_0)$, the optimal value in $s_0$ given the current value of $\theta$. 

On Figures \ref{fig:exp} and \ref{fig:exp2} we observe the evolution of $V_\theta^*(s_0)$ as the number of iterations increases. We observe that $V_\theta^*(s_0)$ converges towards $1-\tau = 0.7$ as the number of iterations increases with oscillations of decreasing amplitude that are due to the ever changing $\theta$ value.  Figures \ref{fig:exp3} and \ref{fig:exp4} show the evolution of the {\em score} as the number of iterations increases. {\em Score} converges towards a value of $0.7$ but inferior. This is due to the exploration of the Q-learning algorithm. Lastly, \ref{fig:exp5} and \ref{fig:exp6} plot the evolution of $\theta$ as the number of iteration increases. The value of $\theta$ converges towards a value of 4 which the one for which  $V_\theta^*(s_0) = 1-\tau=0.7$.  }

\section{Conclusion}\label{sec:conclu}

We have presented an algorithm for learning a policy optimal for the quantile criterion in the reinforcement learning setting, when the MDP has a special structure, which corresponds to repeated episodic decision-making problems.
It is based on stochastic approximation with two timescales \citep{Borkar08}.
Our proposition is experimentally validated on the domain, Who wants to be millionaire.

As future work, it would be interesting to investigate how to choose the learning rate $\alpha_n$ in order to ensure a fast convergence.
Moreover, our approach could be extended to other settings than episodic MDPs.
Besides, it would also be interesting to explore whether gradient-based algorithms could be developed for the optimization of quantiles, based on the fact that a quantile is solution of an optimization problem where the objective function is piecewise linear \citep{Koenker05}.

\newpage
%


\small
\bibliography{biblio160226}

%
%
%
%

\end{document}